\documentclass[5p]{elsarticle}

\journal{Neurocomputing}

\usepackage[utf8]{inputenc}         
\usepackage[T1]{fontenc}            
\usepackage{url}                    
\usepackage{booktabs}               
\usepackage{tabu}                   
\usepackage{amsfonts}               
\usepackage{nicefrac}               
\usepackage{microtype}              

\usepackage{float}                  
\usepackage{subcaption}             
\usepackage{adjustbox}
\usepackage{xspace}
\usepackage{setspace}

\usepackage{amsmath}
\usepackage{amssymb}
\usepackage{mathtools}              
\usepackage[noline,ruled]{algorithm2e}
\SetAlgoInsideSkip{medskip}
\usepackage{enumitem}               
\usepackage{array}
\usepackage{longtable}
\usepackage{multirow}
\usepackage{threeparttable}
\usepackage{makecell}

\usepackage{times}                  
\usepackage{soulutf8}               
\usepackage{hyperref}               

\setuldepth{6.3~~($-$6.3)}          

\DeclareRobustCommand{\ie}{i.e.\@\xspace}

\makeatletter
\DeclareRobustCommand{\etc}{%
    \@ifnextchar{.}%
    {etc}%
    {etc.\@\xspace}%
}

\DeclareRobustCommand\onedot{\futurelet\@let@token\@onedot}
\def\@onedot{\ifx\@let@token.\else.\null\fi\xspace}
\def\etal{\emph{et al}\onedot}

\makeatother

\newcommand{\gph}[2]{\includegraphics[width=#1\linewidth]{#2}}

\newcommand{\bft}[1]{\textbf{#1}}

\newcolumntype{L}[1]{>{\raggedright\arraybackslash}p{#1}}   
\newcolumntype{C}[1]{>{\centering\arraybackslash}p{#1}}     
\newcolumntype{`}{>{\global\let\currentrowstyle\relax}}     
\newcolumntype{~}{>{\currentrowstyle}}                      

\newcommand{\fmtr}[2]{\multirow{#1}{*}{#2}}

\DeclarePairedDelimiter{\floor}{\lfloor \,}{\rfloor}%
\DeclarePairedDelimiter{\abs}{\lvert}{\rvert}%
\DeclarePairedDelimiter{\set}{\{}{\}}%
\newcommand{\BOS}{\texttt{<BOS>}}
\newcommand{\EOS}{\texttt{<EOS>}}

\begin{document}

\begin{frontmatter}

\title{ACORT: A Compact Object Relation Transformer\\for Parameter Efficient Image Captioning}

\author[fsktm]{Jia~Huei~Tan}
\ead{tanjiahuei@siswa.um.edu.my}
\address[fsktm]{CISiP, Faculty of Computer Science and Information Technology, \\ Universiti Malaya, 50603 Kuala Lumpur, Malaysia}

\author[fsktm]{Ying~Hua~Tan}
\ead{tanyinghua@siswa.um.edu.my}

\author[fsktm]{Chee~Seng~Chan\texorpdfstring{\corref{corrauthor}}{*}}
\ead{cs.chan@um.edu.my}
\cortext[corrauthor]{Corresponding author}

\author[engine]{Joon~Huang~Chuah}
\ead{jhchuah@um.edu.my}
\address[engine]{Faculty of Engineering, Universiti Malaya, 50603 Kuala Lumpur, Malaysia}

\begin{abstract}

    Recent research that applies Transformer-based architectures to image captioning has resulted in state-of-the-art image captioning performance, capitalising on the success of Transformers on natural language tasks. Unfortunately, though these models work well, one major flaw is their large model sizes.
    To this end, we present three parameter reduction methods for image captioning Transformers: Radix Encoding, cross-layer parameter sharing, and attention parameter sharing. By combining these methods, our proposed ACORT models have 3.7$\times$ to 21.6$\times$ fewer parameters than the baseline model without compromising test performance. Results on the MS-COCO dataset demonstrate that our ACORT models are competitive against baselines and SOTA approaches, with CIDEr score $\geq$126. Finally, we present qualitative results and ablation studies to demonstrate the efficacy of the proposed changes further. Code and pre-trained models are publicly available at \url{https://github.com/jiahuei/sparse-image-captioning}.

\end{abstract}


\end{frontmatter}



\section{Introduction}
\label{sec: ACORT: Introduction}

Recent years have seen stunning advances in the performance of Deep Neural Networks (DNNs) on sequence modelling tasks, particularly natural language processing, understanding, and generation. One of the main drivers behind these successes is the Transformer model \cite{vaswani2017attention}, which forgoes recurrence in favour of a fully-attentional feed-forward architecture. Besides enabling parallelisation through time during training, Transformers with their residual connections can be stacked to form a multi-layer model. By increasing model depth and width, more powerful models can be trained, and task performance can be improved \cite{sanh2019distilbert}.

Riding on the successes of Transformers, recent research extending Transformer-based architectures to image captioning have resulted in state-of-the-art (SOTA) captioning performance \cite{herdade2019image,cornia2020meshed}. Notably, the Object Relation Transformer (ORT) \cite{herdade2019image} model, which combines the Faster R-CNN detector network from \cite{anderson2018bottom} with a pair of Transformer-based encoder and decoder, improves the SOTA performance by 8 CIDEr\footnote{Consensus-based Image Description Evaluation (CIDEr) is a widely-used metric for measuring caption quality via the level of consensus between ground-truth and generated captions.} score. This is then further improved by \cite{cornia2020meshed} with the Meshed-Memory Transformer.

Unfortunately, whilst these models provide good performance, one main shortcoming is their large model sizes. For example on MS-COCO \cite{lin2014microsoft}, the Soft-Attention model \cite{xu2015show} has only 11.9M parameters, whereas the ORT model has 55.4M parameters (a 4.7$\times$ increase). Furthermore, as the size of datasets grows larger, the vocabulary size increases as well, leading to large input and output embedding matrices which further exacerbates parameter inefficiency. This can impede real-time applications deployment in resource-constrained devices such as mobile and embedded devices. Moreover, large models also consume more memory during training and inference, reducing overall efficiency due to the communication overhead incurred. Finally, recent works on natural language modelling, generation and understanding also suggests that Transformer models are over-parameterised \cite{dabre2019recurrent,mehta2021delight}, a trend that is shared among other DNN models \cite{hinton2015distilling}. 

\begin{figure}[tp]
    \centering
    \gph{0.8}{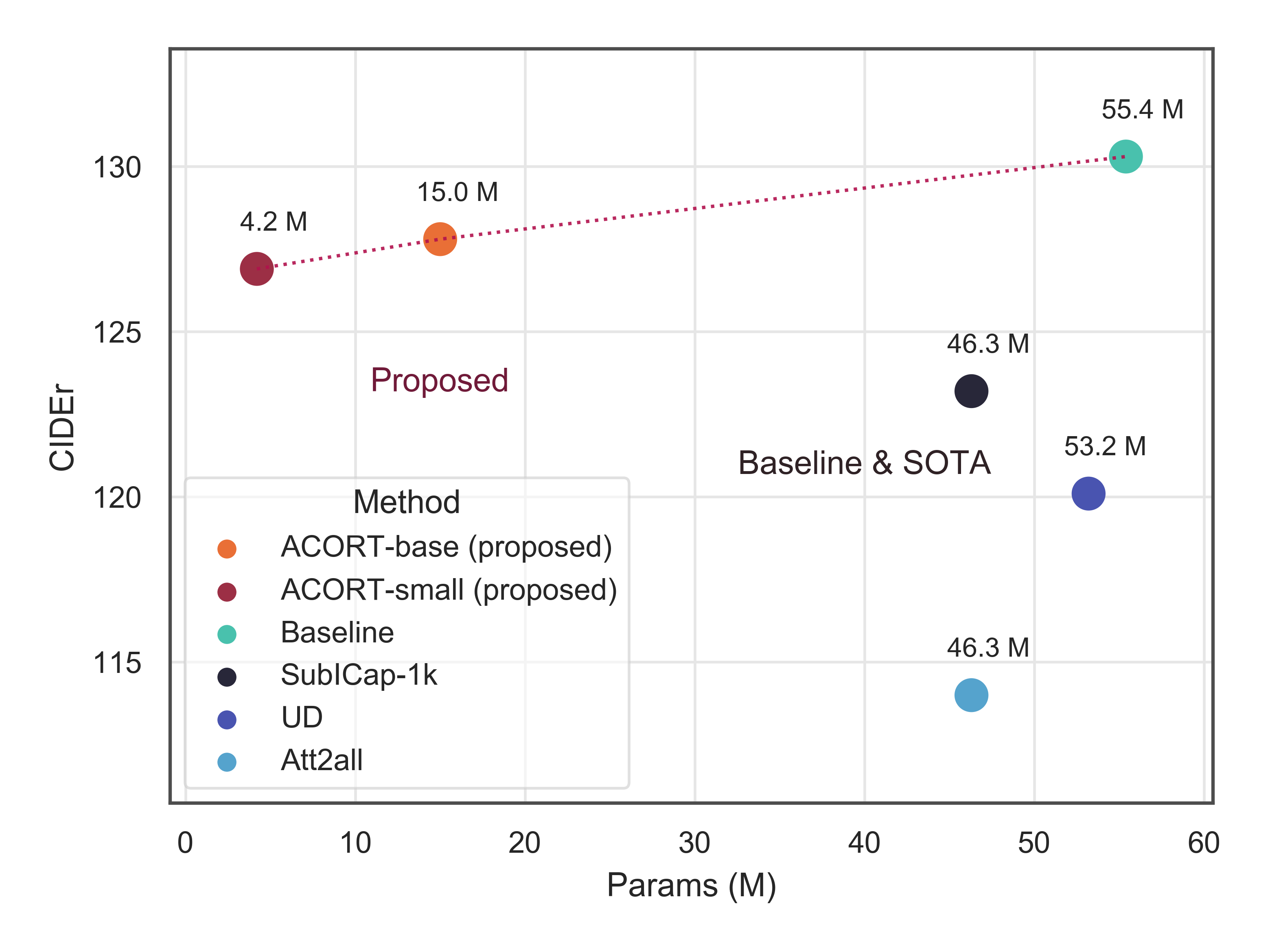}
    \caption{The proposed ACORT models can achieve competitive CIDEr scores on MS-COCO against baseline and state-of-the-art models despite having 3.7$\times$ to 13.2$\times$ fewer parameters. See Table~\ref{table: ACORT: SOTA: Score} for details.}
    \label{fig: Teaser}
\end{figure}

To this end, we address the aforementioned problems in this paper by introducing ACORT -- A Compact Object Relation Transformer architecture for parameter efficient image captioning. By incorporating three parameter reduction methods, ACORT models can achieve performances that are on par with the standard ORT model but with significantly fewer parameters. Firstly, Radix Encoding \cite{tan2019comic} is utilised to drastically reduce the size of the embedding matrices, allowing vocabulary size to grow without affecting model compactness. Secondly, cross-layer parameter sharing is used to decouple the strong correspondence between model depth and model size, allowing more layers to be stacked without increasing parameter count and vice versa. Thirdly, attention parameter sharing is utilised to reduce the parameter count of the multi-head attention module, thereby further improving overall parameter efficiency.

As a result of these optimisations, an ACORT model with the same configuration as ORT has 3.7$\times$ fewer parameters yet maintains the overall performance, while the smallest ACORT model has 21.6$\times$ fewer parameters with minimal performance degradation.
In summary, the core contributions of this work are twofold. Firstly, we propose ACORT, A Compact ORT model with reduced vocabulary size and parameter count (up to 13$\times$ smaller). Secondly, we demonstrate the effectiveness of ACORT on the MS-COCO \cite{lin2014microsoft} dataset. Our experiments show that even the smallest ACORT configuration can reach a CIDEr score $\geq$126, which is comparable to many SOTA approaches (see Fig.~\ref{fig: Teaser} and Sec.~\ref{subsec: ACORT: Experiments: SOTA}).


\section{Related Works}
\label{sec: ACORT: Related Works}

\subsection{Image Captioning}
\label{subsec: ACORT: Related: Image Captioning}

Image captioning is the task of generating descriptive captions given an image. It is a problem that requires scene understanding and natural language generation capabilities. While early works relied on hand-designed templates and rule-based systems, recent works have seen consistent performance improvements driven by the use of DNNs. Here we discuss several notable related works; see \cite{hossain2019comprehensive,liu2019survey} for surveys. 

The end-to-end differentiable encoder-decoder architecture that directly generates a caption given an image was popularised by \cite{karpathy2015deep,vinyals2015show}, and inspired many later works \cite{tan2019phrase,li2020dual}. 
Subsequently, attention was utilised to condition the caption generation process on salient visual features from the image encoder \cite{huang2019attention,pan2020x}. 
Attributes have also been used to inject semantic information into the caption generation process \cite{fang2015captions,wu2017image}. 
Following that, \cite{anderson2018bottom} used an object detector as a form of hard-attention to generate image features from bounding boxes. This framework was then extended by replacing the Long-Short Term Memory (LSTM) decoder with Transformers \cite{vaswani2017attention}. 
Such works include \cite{zhou2018end,yu2019multimodal,cornia2020meshed,yu2021accelerated}, and ORT \cite{herdade2019image}, which is the baseline used in this work. More details on ORT are given in Sec.~\ref{sec: ACORT: ORT}.
At the same time, much work has gone into reinforcement learning, which has enabled the use of non-differentiable caption metrics as optimisation objectives \cite{rennie2017self,chen2018temporal}.
While these methods have been effective in improving SOTA test performance, there has been comparatively little effort put into reducing model size, which is the primary motivation for this paper.

\subsection{Parameter Efficient Transformers}
\label{subsec: ACORT: Related: Efficient Transformers}

Parameter sharing techniques have been utilised to great effect on various Transformer architectures.

\cite{dehghani2019universal} proposed the Universal Transformer, which performs recurrent stacking of layers, effectively performing cross-layer sharing. Such recurrent stacking is also used by RSNMT \cite{dabre2019recurrent}. 
\cite{bai2019deep} suggested deep sequence models produce hidden activations that converge towards some fixed point, and thus proposed the DEQ model that finds these equilibrium points via root-finding to form an infinite depth, weight-tied feedforward network.
Subsequently, \cite{lan2020albert} proposed the ALBERT model, a lightweight Transformer model with cross-layer parameter sharing for pretraining and finetuning on natural language understanding tasks.
Motivated by AlBERT, \cite{lee2020parameter} employed low-rank approximation of Transformer weights via singular value decomposition for video representation learning. More works utilising cross-layer sharing include \cite{sachan2018parameter,reid2021subformer}.
Going a step further, \cite{xia2019tied} performed model-level sharing in which the encoder and the decoder of the Transformer translation model share the same set of weights.
In a similar fashion, the Levenshtein Transformer \cite{gu2019levenshtein} shares the Transformer backbone of three policy classifiers for natural language generation and refinement tasks.

Attention parameter sharing methods were also applied to Transformers.
\cite{xiao2019sharing} shares the self-attention probability maps and the cross-attention outputs across adjacent Transformer layers.
\cite{kitaev2020reformer} proposed the Reformer, which utilised query-key weight sharing to enable the use of locality-sensitive hashing for attention computation.

Despite their remarkable effectiveness, weight sharing strategies have remained largely unexplored for image captioning Transformer models. To this end, we investigate the performance of two different sharing strategies on image captioning, namely cross-layer sharing and attention parameter sharing.

\subsection{Neural Architecture Search, Network Pruning, Quantisation}
\label{subsec: ACORT: Related: NAS}

Methods based on Neural Architecture Search (NAS) have been applied to automatically search for compact architectures that outperform human-designed architectures on performance and efficiency \cite{dong2019network,dong2019searching}. A notable example is the NASNet architecture for image classification, which achieves SOTA accuracy across different levels of computational cost \cite{zoph2018learning}. However, these works have yet to include parameter sharing into their design space. Furthermore, such methods usually involve optimising a separate controller network using reinforcement learning which complicates the training procedure \cite{pham2018efficient,dong2021autohas}. 

On the other hand, techniques such as network pruning, quantisation and distillation can also be used to effectively reduce the number of network parameters \cite{lin2020channel,wen2018learning}. While these methods have been used to compress Transformers \cite{voita2019analyzing,prato2020quantized,ganesh2020compressing}, they are orthogonal to this work. 
In principle, these techniques can be applied on top of our proposed compact model to achieve further parameter and disk space savings.

\subsection{Parameter Efficient Image Captioning}
\label{subsec: ACORT: Related: Efficient Captioning}

\cite{para2017exp} designed a compact image captioning network which combined SqueezeNet \cite{iandola2016squeezenet} and LightRNN \cite{li2016lightrnn} in a Soft-Attention \cite{xu2015show} framework. By using LightRNN with factorised word embeddings, the size of the model is reduced substantially.
Meanwhile, \cite{tan2019comic} proposed the COMIC model, which utilised Radix Encoding, down-projections with weight sharing and multi-head attention to improve the parameter efficiency of Soft-Attention. In the work, Radix Encoding is used to factorise token embeddings across time steps, while weight-tied down-projections are used to reduce the model size further.
\cite{dai2020grow} proposed the H-LSTM cell, which adds extra hidden layers to the vanilla LSTM cell. By doing so, H-LSTM can reduce the number of parameters by requiring fewer external stacked layers. However, their work derives the majority of parameter reduction from the use of the grow-and-prune method \cite{dai2019nest} rather than architectural modifications.
Finally, \cite{sharif2021subicap} proposed the SubICap model, which uses SentencePiece tokenisation \cite{kudo2018sentencepiece} instead of word-level tokenisation. Since SentencePiece performs subword segmentation, its vocabulary size can be made smaller, thus reducing model size.

In this paper, we focus on reducing the number of parameters in Transformers for image captioning. Similar to \cite{sharif2021subicap}, the ORT architecture is chosen as the basis of our work. However, while \cite{sharif2021subicap} only targets the embedding matrices for parameter reduction, our methods perform weight reduction across the entire Transformer. As a result, our proposed models have significantly fewer parameters with comparable test performance. Performance comparisons are given in Sec.~\ref{subsec: ACORT: Experiments: Baseline} and \ref{subsec: ACORT: Experiments: SOTA}.


\section{Object Relation Transformer: Revisited}
\label{sec: ACORT: ORT}

\begin{figure}[t]
    \centering
    \gph{0.8}{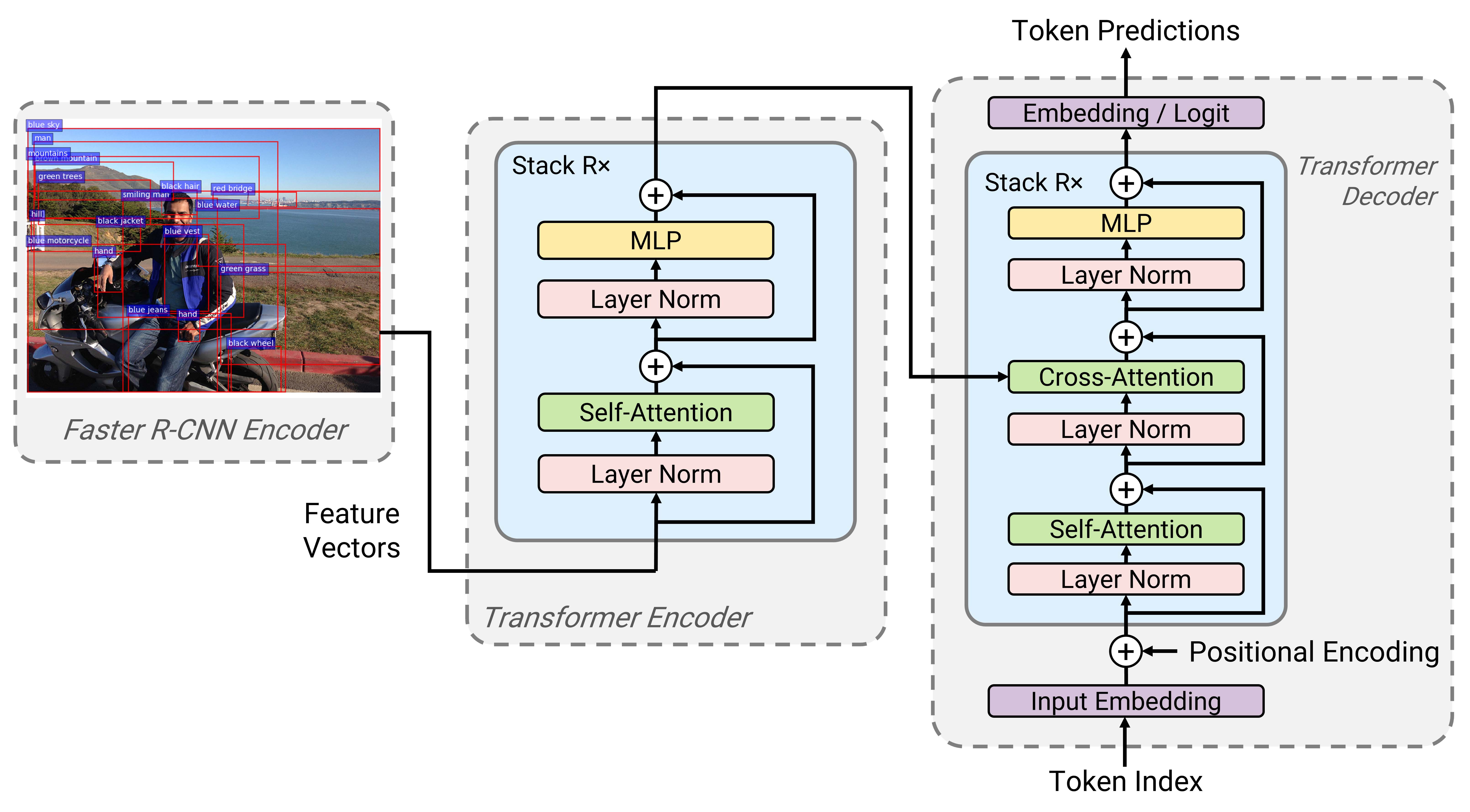}
    \caption{Overview of the Object Relation Transformer (ORT) architecture \cite{herdade2019image}. Each blue box represents a single Transformer layer. Here, the output embedding layer is also known as the logit or pre-softmax projection layer \cite{press2017using}. Figure adapted from \cite{anderson2018bottom,dosovitskiy2021image}.}
    \label{fig: ACORT: ORT}
\end{figure}

Since our proposed ACORT model builds on the \emph{Object Relation Transformer (ORT)} architecture proposed by Herdade \etal{} \cite{herdade2019image}, we present a brief introduction of the ORT model before moving on to our ACORT model. Fig.~\ref{fig: ACORT: ORT} shows an overview of ORT.

The ORT model is an encoder-decoder architecture that extends the Up-Down model of \cite{anderson2018bottom}. The key innovation is the use of geometric attention rather than traditional soft-attention to incorporate spatial relationships between the detected objects into attention weights. The geometric attention module operates by multiplying the standard attention weights between two objects with a learned function of their relative location and scale. These geometric relations are encoded in the form of a spatial coordinate displacement vector along with width and height ratios between the objects.

Moreover, rather than just combining the Faster R-CNN detector network with an LSTM decoder, ORT employs an additional Transformer-based encoder network on top of the detector. The encoder is a 6-layer Transformer network. The R-CNN generated image feature vectors are fed into the encoder Transformer, but they are down-projected from 2,048 channels to 512 channels to fit the Transformer network's dimensionality. Embedded image features are then transformed by a sequence of multi-head scaled dot-product self-attention, multi-layer perceptron (MLP) with ReLU, and layer normalisation \cite{ba2016layer} layers with residual connections \cite{he2016deep} inside the Transformer. These transformed features are then fed into the Transformer decoder's cross-attention module.

The decoder is similar to the encoder, but there are three main differences. First, the decoder's multi-head self-attention performs masked self-attention in order to achieve auto-regressive modelling. Second, the Transformer decoder conducts multi-head cross-attention between the encoder and the decoder in addition to multi-head self-attention. This cross-attention module receives the encoder output features and outputs a convex combination of the features as context vectors. Finally, positional embeddings are used to insert information about the relative positions of the token embeddings that the Transformer decoder receives as input.


\section{ACORT: A Compact Object Relation Transformer}
\label{sec: ACORT: Proposed}

In this section, we present the overall design of ACORT and explain the reasoning behind the components used in ACORT. Quantitative and qualitative comparisons against the standard ORT baseline as well as SOTA approaches are provided in Sec.~\ref{sec: ACORT: Experiments}.

\subsection{Radix Encoding}
\label{subsec: ACORT: Proposed: Radix}

\begin{figure}[t]
    \centering
    \gph{0.8}{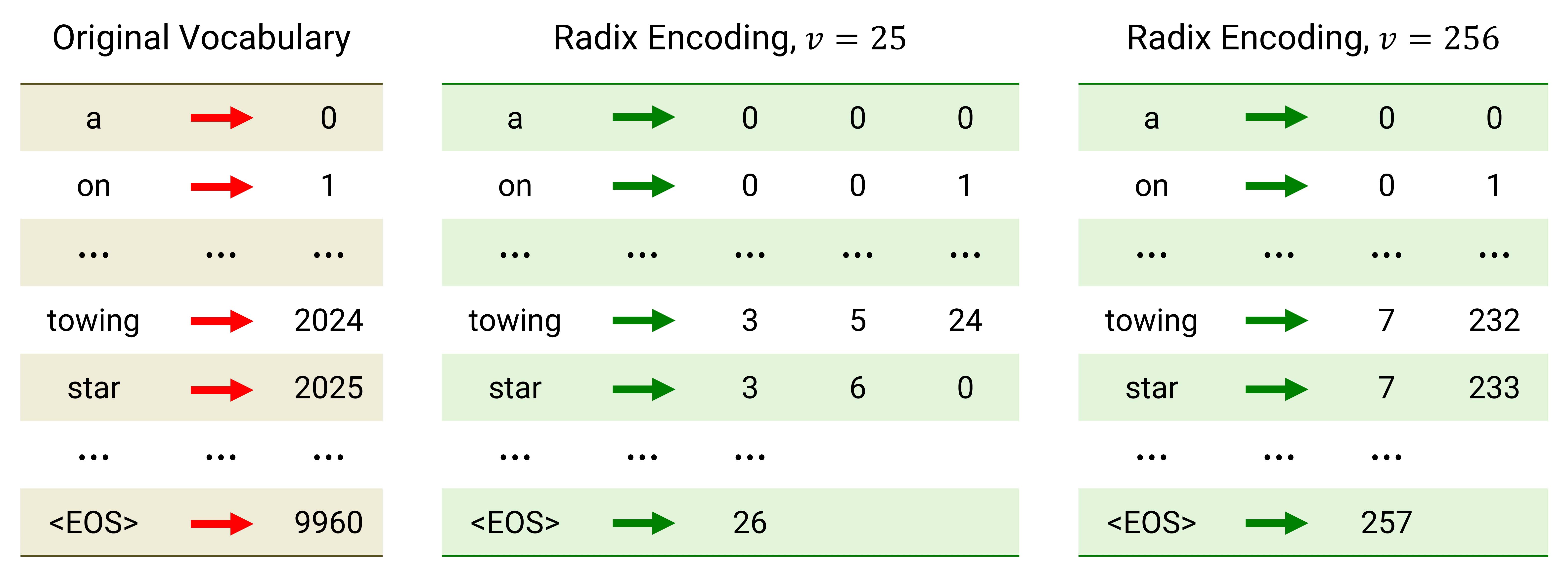}
    \caption{Example of original vocabulary mapping (left), encoded Radix base-25 vocabulary (middle) and encoded Radix base-256 vocabulary (right).}
    \label{fig: ACORT: Radix}
\end{figure}

The original ORT formulation uses a standard architecture whereby a word embedding matrix is used to generate the embedding vector as model input, with a separate output projection matrix used to generate a probability distribution over all the output tokens. On a moderately-sized dataset such as MS-COCO \cite{lin2014microsoft}, the use of word-level tokenisation scheme generates a large vocabulary size and inflated embedding matrices. Typically, these embeddings matrices have sizes that grow in lockstep with the size of the datasets used to train the models. This is due to the fact that that distribution of words in natural language usually follows a Zipfian distribution \cite{baevski2019adaptive}.

To this end, Radix Encoding was proposed in \cite{tan2019comic} as a way to sidestep the aforementioned issue. The method operates by representing each token index $i$ as a combination of indices $\hat{i}$.
Technically, the encoded indices $\hat{i}$ for regular tokens can be computed following Eq.~\eqref{eq: ACORT: Radix regular} to \eqref{eq: ACORT: Radix EOS} below. The word indices are assigned based on word frequency, such that the most common word is assigned to the first index and the least common word is assigned to the last index.

\begin{align}
    \hat{i_j} = \floor{ i \,/\, v ^ j } \text{ mod } v \quad &\text{for } j \in \set{\, 0, \dotsb, d - 1} \text{ and  } i \notin \set{ i_{BOS}, i_{EOS}}, \label{eq: ACORT: Radix regular} \\
    \hat{i} = v \quad &\text{for } i = i_{BOS}, \label{eq: ACORT: Radix BOS} \\
    \hat{i} = v + 1 \quad &\text{for } i = i_{EOS}, \label{eq: ACORT: Radix EOS}
\end{align}

\noindent where $v$ is the Radix base, and $\floor{\cdotp}$ denotes the $floor(\,\cdotp)$ operator converting floating-point values to the nearest integer that is less than or equal to the input. The process is illustrated in Fig.~\ref{fig: ACORT: Radix}. The pseudocode for Radix Encoding is given in Algorithm \ref{algo: ACORT: Radix Encoding}.

Following the equations, token-splitting is performed on all the tokens in the vocabulary, except for the two special tokens: the \BOS{} symbol to start the caption generation process, and the \EOS{} symbol to signal the end of the generation process. Doing so allows the original vocabulary $V_o$ of size $v^d$ to be compactly encoded into a new vocabulary $V_e$ of size $v + 2$, where $d$ is the number of Radix tokens required to encode each original token. 
Note that $d$ and $v$ are inversely proportional to each other, where for a given vocabulary size $\abs{V_o}$, a smaller Radix base $v$ will lead to a larger $d$. In practice, $d = 2$ is able to provide good performance while maintaining model compactness.

By employing the encoding scheme on top of the word-level tokenisation, we can compress the original embedding matrices from a size of $v^d \times r$ to $(v + 2) \times r$. This leads to an exponential reduction in embedding sizes at a factor of $d$, in which $\abs{V_e} \ll \abs{V_o}$. There is also no need to alter model code as one can simply run inference using greedy or beam search (or other inference methods) as usual and apply post-processing on the output tokens. 

\begin{algorithm}
    \setstretch{1.5}
    \caption{Radix Encoding}
    \label{algo: ACORT: Radix Encoding}
    \SetKwInput{KwReq}{Require}
    \SetKw{Let}{Let}
    \SetKwComment{Comment}{$\Rightarrow$ }{}
    \SetCommentSty{}    
    \KwIn{Training corpus $D$, Radix base $v$}
    \KwOut{Word-to-radix mapping $V_e$}

    $count \gets$ List of words and its corresponding frequency in $D$ \;
    $count \gets$ Sort $count$ by word frequency in descending order \;
    $V_o \gets$ Assign an index to each word in $count$
    \Comment*{The original word vocabulary}
    
    $V_e \gets$ Empty dictionary
    \Comment*{The Radix vocabulary}
    \ForEach{(word, $i$) in $V_o$}{
        Compute $(\hat{i_0}, \dotsb, \hat{i_j})$ from $i$ and $v$ using Eq.~\eqref{eq: ACORT: Radix regular} \;
        $V_e[word] \gets (\hat{i_0}, \dotsb, \hat{i_j})$ \;
    }
    $V_e[BOS] \gets v$
    \Comment*{Refer to Eq.~\eqref{eq: ACORT: Radix BOS}}
    $V_e[EOS] \gets v + 1$
    \Comment*{Refer to Eq.~\eqref{eq: ACORT: Radix EOS}}
    \Return{$V_e$}
\end{algorithm}

\subsection{Cross-Layer Sharing}
\label{subsec: ACORT: Proposed: Cross-layer}

\begin{figure}[t]
    \centering
    \gph{0.8}{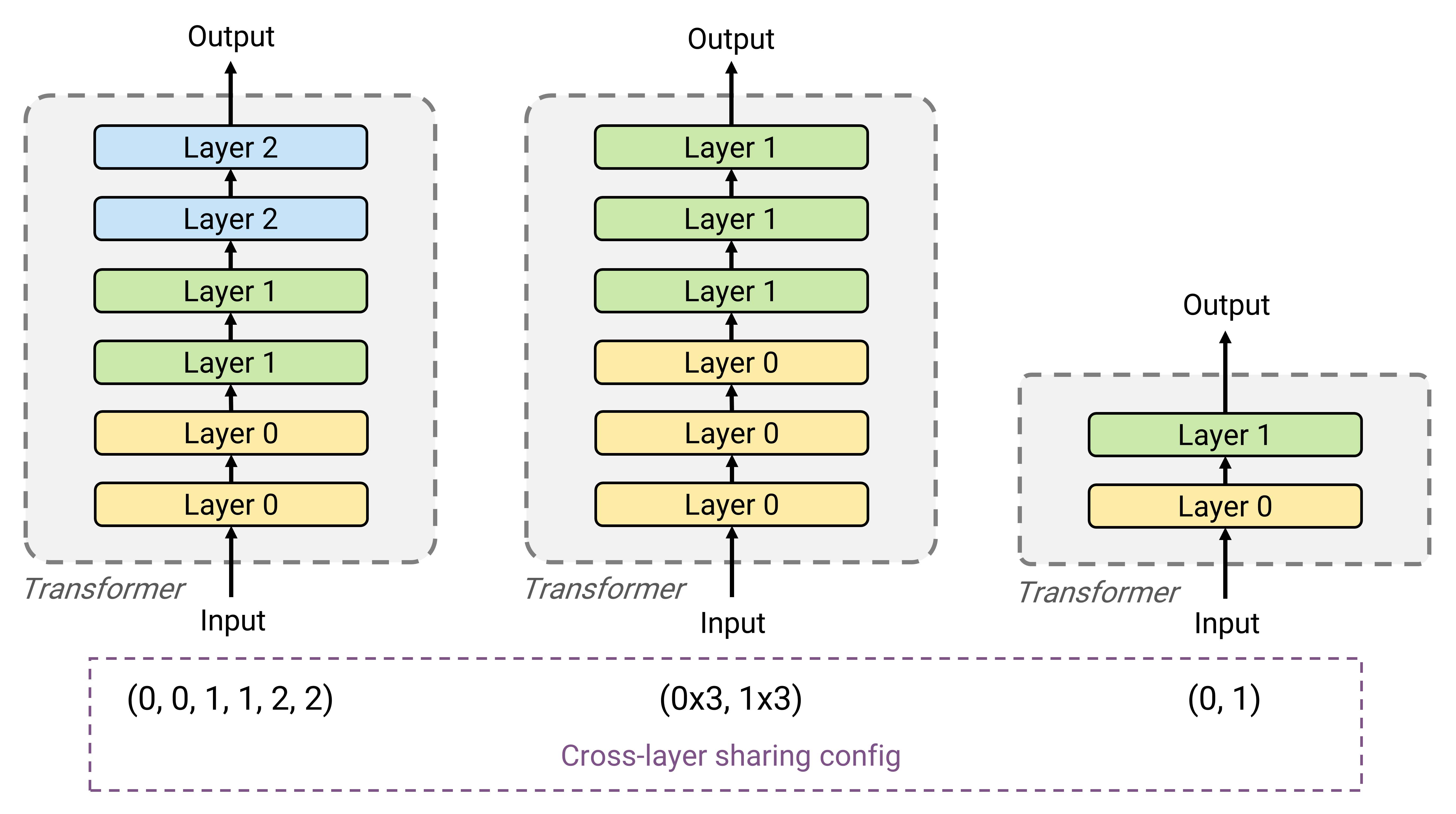}
    \caption{
    Example configurations of cross-layer sharing in a Transformer. 
    \texttt{(0,0,1,1,2,2)} means that there are 6 layers in total and sharing is applied every two layers. 
    \texttt{(0x3,1x3)} is similar except sharing is applied every three layers. 
    \texttt{(0,1)} means that there are 2 layers in total and no sharing is applied.
    }
    \label{fig: ACORT: Cross-layer}
\end{figure}

In addition to Radix Encoding, cross-layer sharing is utilised as another effective method for parameter reduction in ACORT. A Transformer with cross-layer sharing applied will have a subset or all of its layers share parameters. In the case where the number of layers is larger than the number of independent layers, this can lead to significant parameter reductions. For clarity, the layer configurations are denoted as follows: A 6-layer Transformer with 3 independent layers (\ie{} 3 layers shared across 2 layers each) is denoted as \texttt{(0,0,1,1,2,2)}, a 6-layer Transformer with 2 independent layers is denoted as \texttt{(0,0,0,1,1,1)} or \texttt{(0x3,1x3)}, and a 2-layer Transformer with 2 independent layers is denoted as \texttt{(0,1)}. These configurations are illustrated in Fig.~\ref{fig: ACORT: Cross-layer}.

As outlined in Sec.~\ref{subsec: ACORT: Related: Efficient Transformers}, there are many ways to share Transformer parameters across its layers. A notable example is the ALBERT model, which shares all parameters across its layers \cite{lan2020albert}. This means that a 6-layer ALBERT will have a layer-sharing configuration of \texttt{(0x6)}. While this aggressive sharing scheme yields a model that is very lightweight, it is not the ideal configuration for image captioning (see Table~\ref{table: ACORT: Ablation: Layer share}).

In this paper, we explore the application of cross-layer sharing in a group-wise fashion, such that there is more than one independent layer in the Transformer. Specifically, we tested various layer-sharing configurations with the number of independent layers ranging from one to five. Empirical result in Sec.~\ref{subsec: ACORT: Experiments: Ablations} shows that a model with two independent layers can provide good performance and parameter count reduction.

\subsection{Attention Parameter Sharing}
\label{subsec: ACORT: Proposed: Attention}

\begin{figure}[t]
    \centering
    \gph{0.8}{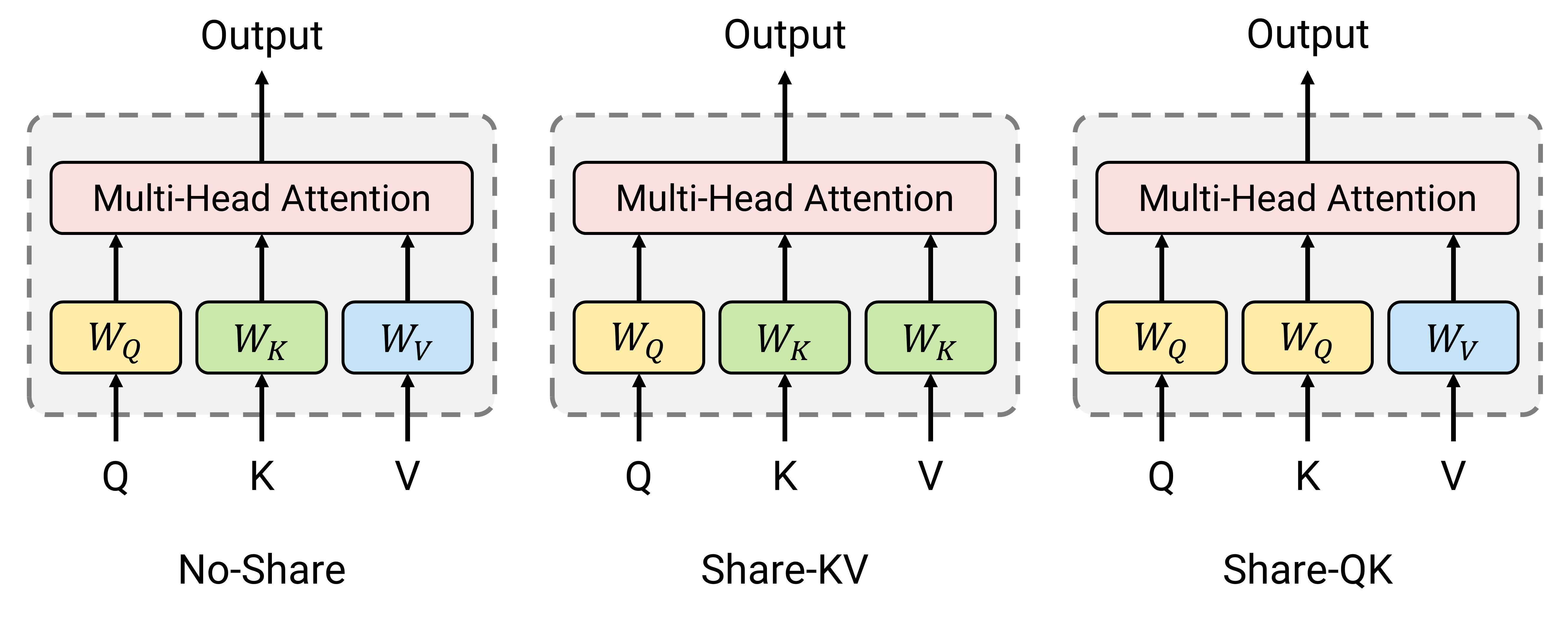}
    \caption{
    Example of attention parameter sharing. 
    \textit{No-Share} is the standard baseline attention. 
    \textit{Share-KV} shares $W_K$ and $W_V$, while 
    \textit{Share-QK} shares $W_Q$ and $W_K$. 
    }
    \label{fig: ACORT: Attention}
\end{figure}

While cross-layer sharing is an effective inter-layer parameter sharing strategy, further weight savings can be derived by leveraging intra-layer parameter sharing. Specifically, each of the multi-head scaled dot-product attention modules in a Transformer layer contains three weight matrices $W_Q, W_K, W_V$ that can be potentially shared. Fig.~\ref{fig: ACORT: Attention} shows two types of parameter sharing scheme explored in this paper: \textit{Share-KV} and \textit{Share-QK}.

\begin{figure}[t]
    \centering
    \gph{0.8}{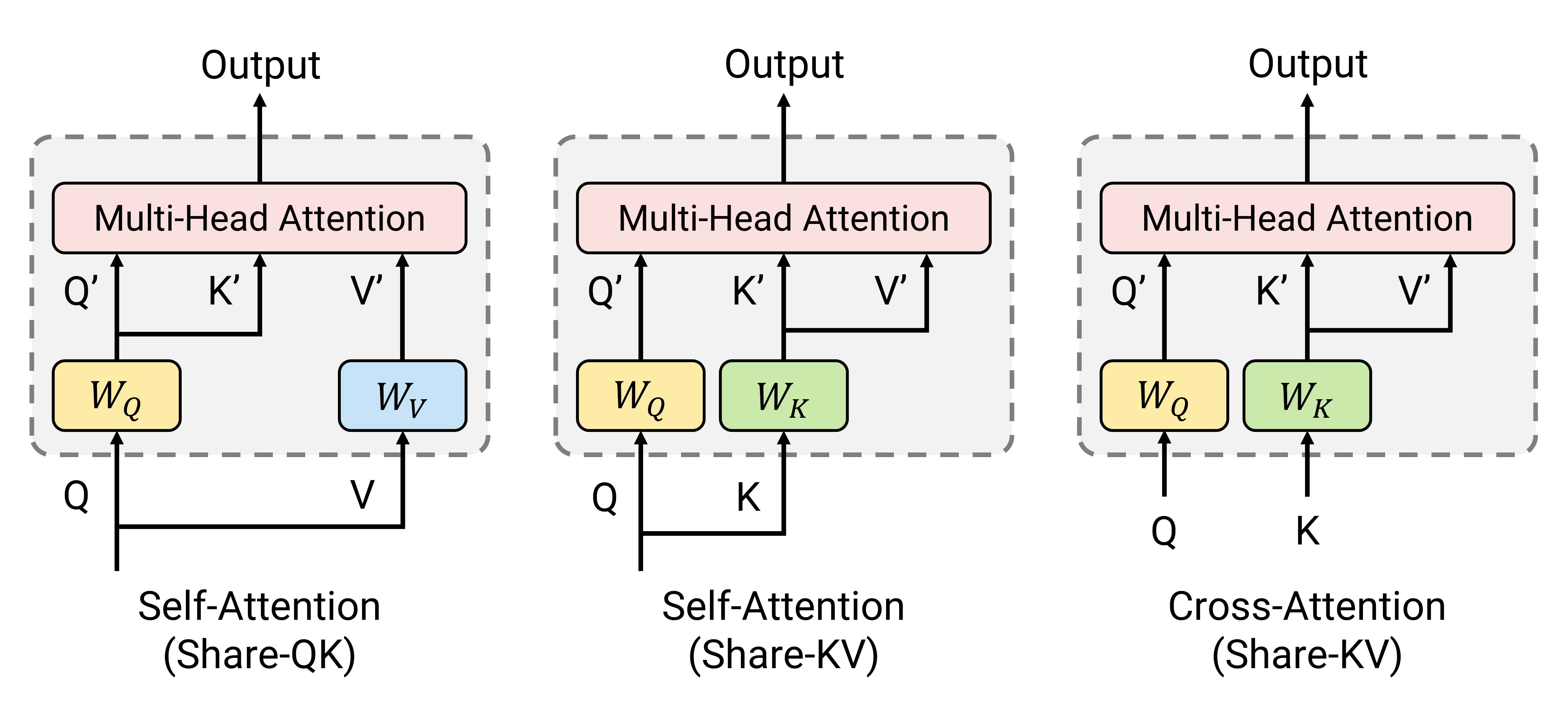}
    \caption{When attention parameter sharing is used, some projection computations can be skipped by reusing the output of the shared projection layer.}
    \label{fig: ACORT: Attention skip}
\end{figure}

In a nutshell, \textit{Share-QK} operates by sharing the parameters of the weight matrices $W_Q$ and $W_K$. Likewise, \textit{Share-KV} performs parameter sharing on the weight matrices $W_K$ and $W_V$. In addition to reducing learnable parameters, these weight sharing schemes can reduce computation cost by allowing some linear projection operations to be skipped. This is achieved by reusing the output from the shared intermediate projection layer, illustrated in Fig.~\ref{fig: ACORT: Attention skip}. 

In self-attention, all three inputs to the attention module -- query, key, value -- originates from the same source tensor. In other words, the inputs to self-attention are identical among each other ($Q \equiv K \equiv V$). Thus for self-attention, both \textit{Share-QK} and \textit{Share-KV} allow for intermediate projection reuse. However, this is not the case for cross-attention where query comes from the decoder while key and value come from the encoder. Hence for cross-attention, only \textit{Share-KV} is able to skip the computation of value projection by directly reusing the projected key (key-value reuse).

\subsection{Final Proposed Setup}
\label{subsec: ACORT: Proposed: Final}

\begin{table}[t]
    \caption{The configurations of the ORT and ACORT models, unless otherwise stated.}
    \label{table: ACORT: Proposed: Setup}
    \centering
    \begin{adjustbox}{max width=\linewidth}
    \begin{tabular}{ `l ~l ~c ~c ~c ~c ~c ~c }
        \toprule
        Approach & & \makecell{Params\\(M)} & \makecell{Radix\\base $v$} & \makecell{Layer\\config} 
        & \makecell{Att. param\\sharing} & \makecell{Hidden\\size} & \makecell{MLP\\size} \\
        \midrule
\multirow{5}{4em}{ORT (baseline)} & base & 55.4 &               &                    &              & 512         & 2,048    \\
                        & small  & 16.7       & -               & (0, 1, 2, 3, 4, 5) & No-Share     & 256         & 1,024    \\
                        & xsmall & 4.1        &                 &                    &              & 104         & 416      \\
                        \cmidrule(lr){2-8}
                        & base-4 & 40.7       & \fmtr{2}{-}     & (0, 1, 2, 3) & \fmtr{2}{No-Share} & \fmtr{2}{512} & \fmtr{2}{2,048} \\
                        & base-2 & 26.0       &                 & (0, 1)             &              &             &          \\
        \midrule
        \fmtr{4}{ACORT} & base   & 15.0       & \fmtr{2}{768}   & (0x3, 1x3)   & \fmtr{2}{Share-KV} & \fmtr{2}{512} & \fmtr{2}{2,048} \\
                        & base-AL & 8.4       &                 & (0x6)              &              &             &          \\
                        \cmidrule(lr){2-8}
                        & small  & 4.2        & \fmtr{2}{768}   & (0x3, 1x3)   & \fmtr{2}{Share-KV} & \fmtr{2}{256} & \fmtr{2}{1,024} \\
                        & xsmall & 2.6        &                 & (0x2)              &              &             &         \\
        \bottomrule
    \end{tabular}
    \end{adjustbox}
\end{table}

Our proposed model ACORT incorporates all of the aforementioned parameter reduction strategies, namely Radix Encoding, cross-layer parameter sharing, and attention parameter sharing. The differences in model size and setup compared to the standard ORT baseline are presented in Table~\ref{table: ACORT: Proposed: Setup}.

\textbf{Radix Encoding:} ACORT utilises a base number of $v = 768$, which provided a good balance of performance and parameter efficiency (see Sec.~\ref{subsec: ACORT: Experiments: SOTA}). Other Radix base values are also explored in Table~\ref{table: ACORT: Ablation: Radix}, from $v = 256$ up to $v = 1024$. 

\textbf{Cross-layer sharing:} ACORT uses \texttt{(0x3,1x3)} layer-sharing for the \textit{-base} and \textit{-small} configurations. For the \textit{-xsmall} configuration, \texttt{(0x2)} layer-sharing is used to yield maximum parameter reduction. We also evaluated the performance of an ALBERT style layer-sharing scheme \texttt{(0x6)}, denoted as the \textit{ACORT-base-AL} configuration.

\textbf{Attention parameter sharing:} ACORT uses \textit{Share-KV} as the attention parameter sharing method, as it shows good performance compared to both \textit{Share-QK} and the baseline (see Table~\ref{table: ACORT: Ablation: Attention}). Moreover, it has the slight advantage of allowing key-value reuse for both self-attention and cross-attention.

As a result of these optimisations, the ACORT-base configuration has 3.7$\times$ fewer parameters than that of ORT-base (55.4M $\rightarrow$ 15.0M). The ACORT-small configuration has 13.2$\times$ fewer parameters than ORT-base (55.4M $\rightarrow$ 4.2M), and 4.0$\times$ fewer parameters than ORT-small (16.7M $\rightarrow$ 4.2M). The smallest ACORT configuration has 21.6$\times$ fewer parameters than ORT-base (55.44M $\rightarrow$ 2.57M), yet it is able to reach a CIDEr score of 126, which is comparable to many SOTA approaches (see Sec.~\ref{subsec: ACORT: Experiments: SOTA}). Meanwhile, even though the ORT \textit{-small} and \textit{-xsmall} configurations are compact, their performances are less ideal.

In terms of baseline models, both the \textit{-small} and \textit{-xsmall} configurations have the same number of layers as the \textit{-base} model, but with narrower layers. In addition, we also included the \textit{-base-2} and \textit{-base-4} configurations with the same width as the \textit{-base} model, but with fewer layers. Performance comparison between baselines and ACORT is given in Sec.\ref{subsec: ACORT: Experiments: Baseline}.


\section{Experiments}
\label{sec: ACORT: Experiments}

In this section, we present empirical results on the performance of our ACORT model, and provide comparisons against both baseline and SOTA approaches. We also provide extensive ablation studies on the contribution of each component in ACORT.

\subsection{Experiment Setup}
\label{subsec: ACORT: Experiments: Setup}

Our experiment setup closely follows ORT \cite{herdade2019image} to allow for meaningful quantitative comparisons. 

\bft{Implementation and hyperparameters:} We reuse the public implementations published by the authors\footnote{\url{https://github.com/yahoo/object_relation_transformer}} for both ACORT and baseline models.
Following \cite{herdade2019image}, Adam is utilised as the optimiser, with the ``Noam'' learning rate scheduler for cross-entropy training and the step learning rate schedule for SCST training \cite{rennie2017self}.

Our SCST training utilised the recently proposed method by \cite{cornia2020meshed,luo2020better} for performing action space sampling and computing the reward baseline. Random search is used to generate multiple sampled captions for each image, and the baseline for each image is set to be the average of the caption scores.

Inference is performed using beam search without length normalisation, using the model checkpoint with the highest validation CIDEr score. Following \cite{herdade2019image}, test set performance is obtained using beam size of 2 after cross-entropy optimisation; and beam size of 5 after SCST optimisation.

\bft{Datasets and metrics:} Experiments are performed on MS-COCO \cite{lin2014microsoft}, a popular English captioning dataset for benchmarking. 
The ``Karpathy'' split \cite{karpathy2015deep} is utilised, which assigns 5,000 images for validation, 5,000 for testing and the rest for training. Pre-processing of captions is done following \cite{tan2019comic}.
Evaluation scores are obtained using the publicly available MS-COCO evaluation toolkit\footnote{\url{https://github.com/salaniz/pycocoevalcap}}, which computes BLEU, METEOR, ROUGE-L, CIDEr and SPICE (B, M, R, C, S).

\subsection{Baseline Comparison}
\label{subsec: ACORT: Experiments: Baseline}

\subsubsection{Test performance}
\label{subsubsec: ACORT: Baseline: Test performance}

\begin{table}[t]
    \caption{Test set performance comparison against the baseline ORT models. ACORT models provide competitive test performance with significantly fewer parameters when compared to baseline models.}
    \label{table: ACORT: Baseline: Score}
    \centering
    \begin{adjustbox}{max width=\linewidth}
    \begin{tabular}{ `l ~l ~c ~c ~c ~c ~c ~c ~c ~c ~c }
        \toprule
        \multicolumn{2}{c}{\fmtr{2.35}{Approach}} & \multicolumn{1}{c}{\fmtr{2.35}{\makecell{Params\\(M)}}} 
        & \multicolumn{6}{c}{MS-COCO test scores} \\
        \cmidrule(lr){4-9}
                                            &           &              & B-1  & B-4  & M    & R    & C     & S    \\
        \midrule
        \multirow{5}{4em}{ORT (baseline)}   & base      & 55.4         & 76.1 & 35.2 & 27.9 & 56.5 & 114.7 & 21.0 \\
                                            & base-4    & 40.7         & 76.4 & 35.3 & 27.8 & 56.4 & 114.0 & 21.1 \\
                                            & base-2    & 26.0         & 76.7 & 35.9 & 27.8 & 56.6 & 114.8 & 21.2 \\
                                            & small     & 16.7         & 76.1 & 35.1 & 27.5 & 56.2 & 113.0 & 20.6 \\
                                            & xsmall    & 4.1          & 74.9 & 32.7 & 26.2 & 54.9 & 104.4 & 19.1 \\
        \midrule
        \multirow{4}{*}{ACORT}              & base      & 15.0         & 76.6 & 35.9 & 28.1 & 56.9 & 116.0 & 21.4 \\
                                            & base-AL   & 8.4          & 75.6 & 33.8 & 27.2 & 55.7 & 110.0 & 20.8 \\
                                            & small     & 4.2          & 76.0 & 34.6 & 27.4 & 56.0 & 112.5 & 20.7 \\
                                            & xsmall    & 2.6          & 75.1 & 33.7 & 27.1 & 55.6 & 108.7 & 20.3 \\
        \bottomrule
    \end{tabular}
    \end{adjustbox}
\end{table}

Table~\ref{table: ACORT: Baseline: Score} shows the comparison between model size and the metric scores of our model against the baseline after cross-entropy (teacher-forcing) training. The configurations presented here are detailed in Sec.~\ref{subsec: ACORT: Proposed: Final}.

Overall, the performance of ACORT is on par with the various baseline configurations. Comparing the \textit{-base} configurations, our ACORT-base model is able to slightly outperform ORT-base, ORT-base-4 and ORT-base-2 models, despite having 42\% to 73\% fewer parameters. It achieves relative improvements of 2.0\% on BLEU-4, 1.1\% on CIDEr, and 1.9\% on SPICE compared to ORT-base. Interestingly, ORT-base-2 outperforms both ORT-base and ORT-base-4, suggesting that the original configuration might be overparameterised and could benefit from the regularisation effects of simply having fewer parameters.

In addition, ACORT-small can provide competitive test performance even with a large 92.4\% reduction in parameter count (13$\times$). Compared to ORT-base, its scores dropped by merely 1.7\% on BLEU-4, 1.9\% on CIDEr, and 1.4\% on SPICE. When compared to ORT-xsmall with a similar parameter count, ACORT-small clearly outperforms it, with relative improvements of 5.4\% on BLEU-4, 4.3\% on CIDEr, and 1.9\% on SPICE. At the same time, ACORT-xsmall with only 2.6 M parameters (38.0\% reduction) also outperforms ORT-xsmall. While a considerable score gap exists between ACORT-xsmall and ORT-base, the CIDEr gap can be reduced significantly through the SCST training procedure (see Sec.~\ref{subsec: ACORT: Experiments: SOTA}). On the other hand, ACORT-base-AL with only one independent layer underperforms compared to both ACORT-base and ACORT-small, with score gaps of $-$2.3\% on BLEU-4, $-$2.2\% on CIDEr, and $+$0.5\% on SPICE compared to ACORT-small.

One notable observation from our experiments is the training instability of the ORT-xsmall configuration. The model tends to diverge during training, and we had to repeat its training run twice to get a model that did not diverge early in the process. Other than that, all the other configurations had stable training runs, including the ACORT models.

\subsubsection{Caption uniqueness and length}
\label{subsubsec: ACORT: Baseline: Caption uniqueness}

\begin{table}[t]
    \caption{Uniqueness and average length of the generated captions on MS-COCO. Captions generated by ACORT contain 2.0\% to 4.6\% more unseen sentences that are $\sim$0.2 words longer.}
    \label{table: ACORT: Baseline: Caption stats}
    \centering
    \begin{adjustbox}{max width=\linewidth}
    \begin{threeparttable}
    \begin{tabular}{ `l ~l ~c ~c }
        \toprule
        \multicolumn{2}{c}{Approach}        & Unique (\%) \tnote{a}     & Average word count \\
        \midrule
        \multirow{5}{4em}{ORT (baseline)}   & base      & 61.2          & 9.52           \\
                                            & base-4    & 60.1          & 9.44           \\
                                            & base-2    & 60.6          & 9.38           \\
                                            & small     & 60.4          & 9.39           \\
                                            & xsmall    & 57.6          & 9.23           \\
        \midrule
        \multirow{4}{*}{ACORT}              & base      & 63.2          & 9.59           \\
                                            & base-AL   & 71.7          & 9.73           \\
                                            & small     & 62.2          & 9.44           \\
                                            & xsmall    & 62.8          & 9.53           \\
        \bottomrule
    \end{tabular}
    \begin{tablenotes}
        \item[a] Percentage of generated captions not found in training set.
    \end{tablenotes}
    \end{threeparttable}
    \end{adjustbox}
\end{table}

To further assess the quality of captions generated by the models, we calculated the captions' uniqueness and average word count. Results are given in Table~\ref{table: ACORT: Baseline: Caption stats}. Once again, the ACORT models were able to produce good performance. In terms of both caption uniqueness and length, ACORT is able to outperform the baselines. ACORT-generated captions contain 1.0\% to 14.1\% more unseen sentences, and the average lengths are around 0.2 words longer.

\subsubsection{Training and inference cost}
\label{subsubsec: ACORT: Baseline: Training and inference cost}

\begin{table}[t]
    \caption{Training and inference cost comparison against the baseline ORT models. ACORT models consume less GPU memory during training with 20\% to 432\% faster training speeds compared to ORT-base.}
    \label{table: ACORT: Baseline: Speedup}
    \centering
    \begin{adjustbox}{max width=\linewidth}
    \begin{tabular}{ `l ~l ~c ~c ~c ~c ~c }
        \toprule
        \multicolumn{2}{c}{\fmtr{2.35}{Approach}} & 
        \multicolumn{1}{c}{\fmtr{2.35}{\makecell{Params\\(M)}}} & 
        \multicolumn{2}{c}{GPU Memory ($\downarrow$)} & 
        \multicolumn{2}{c}{Relative Speedup ($\uparrow$)} \\
        \cmidrule(lr){4-5} \cmidrule(lr){6-7}
                                            &         &      & GB   & Relative & Training & Inference \\
        \midrule
        \multirow{5}{4em}{ORT (baseline)}   & base    & 55.4 & 2.54 & 1.0$\times$ & 1.0$\times$ & 1.0$\times$ \\
                                            & base-4  & 40.7 & 2.00 & 0.8$\times$ & 1.4$\times$ & 1.1$\times$ \\
                                            & base-2  & 26.0 & 1.22 & 0.5$\times$ & 2.5$\times$ & 1.3$\times$ \\
                                            & small   & 16.7 & 1.53 & 0.6$\times$ & 1.1$\times$ & 0.9$\times$ \\
                                            & xsmall  & 4.1  & 0.87 & 0.3$\times$ & 1.1$\times$ & 0.9$\times$ \\
        \midrule
        \multirow{4}{*}{ACORT}              & base    & 15.0 & 2.36 & 0.9$\times$ & 1.2$\times$ & 0.8$\times$ \\
                                            & base-AL & 8.4  & 2.29 & 0.9$\times$ & 1.3$\times$ & 1.1$\times$ \\
                                            & small   & 4.2  & 1.50 & 0.6$\times$ & 1.4$\times$ & 0.8$\times$ \\
                                            & xsmall  & 2.6  & 0.57 & 0.2$\times$ & 4.3$\times$ & 1.2$\times$ \\
        \bottomrule
    \end{tabular}
    \end{adjustbox}
\end{table}

We examine the training and inference costs of our ACORT models as well as the baseline models. We measured the GPU memory required for the training process using figures reported by \textit{torch.cuda.memory\_reserved()} at the end of the first 1,000 training steps. Similarly, the training speedups were computed based on the time taken to perform the first 1,000 model updates. Inference speedups were calculated based on the time taken to generate captions on the entire test set with a batch size of 50 and a beam size of 2, following the setup described in Sec.~\ref{subsec: ACORT: Experiments: Setup}. All results were obtained using a Titan Xp GPU.

Table~\ref{table: ACORT: Baseline: Speedup} shows that the ACORT models consume less GPU memory during training due to their lower parameter counts. This is especially true for ACORT-xsmall, which is both extremely compact with 2.6 M parameters and lightweight with only 2 layers in total. These are also reflected in the training speedups enjoyed by the ACORT models, with speedups ranging from 1.2$\times$ to 4.3$\times$. In terms of inference speeds, the ACORT models suffered slightly at 0.8$\times$ speedup compared to the baselines. This can be attributed to the increased sequence length incurred by the use of Radix Encoding. Further discussion is provided in Sec.~\ref{sec: ACORT: Limitations}. However, we note that our ACORT-xsmall model can still provide a 1.2$\times$ inference speedup compared to ORT-base, with minimal performance degradation after SCST optimisation is completed (see Table~\ref{table: ACORT: SOTA: Score}).

All in all, these results show that the proposed modifications are effective at reducing model parameter count while maintaining test performance.

\subsection{State-of-the-Art Comparison}
\label{subsec: ACORT: Experiments: SOTA}

\begin{table}[t]
    \caption{Test set performance of the proposed model ACORT along with SOTA methods on MS-COCO. ACORT models provide good performance-to-size ratio, with parameter reductions of up to 18$\times$ when compared to SubICap-1k.}
    \label{table: ACORT: SOTA: Score}
    \centering
    \begin{adjustbox}{max width=\linewidth}
    \begin{threeparttable}
    \begin{tabular}{ `l ~c ~c ~c ~c ~c ~c ~c ~c ~c ~c }
        \toprule
        \multicolumn{2}{c}{\fmtr{2.35}{Approach}} 
        & \fmtr{2.35}{\makecell{Params\\(M) \tnote{a}}} & \multicolumn{8}{c}{MS-COCO test scores (single-model)}            \\
                                                \cmidrule(lr){4-11}
                                             &        &             & B-1  & B-2  & B-3  & B-4  & M    & R    & C     & S    \\
        \midrule
\multicolumn{2}{l}{Att2all \cite{rennie2017self}}  & 46.3 \tnote{b} & -    & -    & -    & 34.2 & 26.7 & 55.7 & 114.0 & -    \\
        UD \cite{anderson2018bottom}         &     & 53.2 \tnote{b} & 79.8 & -    & -    & 36.3 & 27.7 & 56.9 & 120.1 & 21.4 \\
        ORT \cite{herdade2019image}          &     & 55.4 \tnote{b} & 80.5 & -    & -    & 38.6 & 28.7 & 58.4 & 128.3 & 22.6 \\
        AoANet \cite{huang2019attention}    &       & -             & 80.2 & -    & -    & 38.9 & 29.2 & 58.8 & 129.8 & 22.4 \\
        M2 \cite{cornia2020meshed}          &       & -             & 80.8 & -    & -    & 39.1 & 29.2 & 58.6 & 131.2 & 22.6 \\
        X-Transformer \cite{pan2020x}       &       & -             & 80.9 & 65.8 & 51.5 & 39.7 & 29.5 & 59.1 & 132.8 & 23.4 \\
        \midrule
\multicolumn{2}{l}{LightRNN \cite{para2017exp} \tnote{c}} & 11.8    & 67.5 & 46.5 & 32.1 & 22.6 & 22.0 & -    & -     & -    \\
\multicolumn{2}{l}{H-LSTM \cite{dai2020grow} \tnote{c}} & -         & 71.9 & -    & -    & -    & -    & -    & 95.4  & -    \\
\multicolumn{2}{l}{ComIC-256 \cite{tan2019comic}}       & 4.0       & 75.3 & -    & -    & 34.4 & -    & -    & 105.0 & 19.0 \\
\multicolumn{2}{l}{SubICap-1k \cite{sharif2021subicap}} & 46.3      & 79.5 & -    & -    & 37.1 & 29.8 & 58.2 & 123.2 & 23.0 \\
        \midrule
        \multirow{3}{*}{ACORT}               & base   & 15.0        & 79.3 & 64.1 & 49.9 & 38.2 & 28.4 & 57.8 & 127.8 & 21.6 \\
                                             & small  & 4.2         & 79.7 & 64.4 & 50.2 & 38.5 & 28.1 & 57.6 & 126.9 & 21.3 \\
                                             & xsmall & 2.6         & 79.0 & 63.5 & 49.3 & 37.7 & 28.0 & 57.3 & 126.0 & 21.2 \\
        \bottomrule
    \end{tabular}
    \begin{tablenotes}
        \item[a] Excluding image encoder.
        \item[b] Calculated based on reimplementation.
        \item[c] Teacher-forcing training only.
    \end{tablenotes}
    \end{threeparttable}
    \end{adjustbox}
\end{table}

Table~\ref{table: ACORT: SOTA: Score} demonstrates the test set performance of ACORT models compared to state-of-the-art approaches. The compact image captioning models are separated into the second group in the table.

Compared to compact image captioning models, ACORT models provide better performance while having smaller model sizes. ACORT is able to outperform all of the existing approaches on most metrics, with the lone exception of \textit{SubICap-1k}. By comparing ACORT against COMIC-256, the performance improvements derived from better image features (Faster R-CNN) and sequence decoder (Transformer) are apparent: ACORT-xsmall achieves $+$9.6\% on BLEU-4, $+$20.0\% on CIDEr, and $+$11.6\% on SPICE. At the same time, ACORT-xsmall is 36\% smaller than COMIC-256 (1.47 M decrease in parameters).

Against the ORT-based \textit{SubICap-1k}, ACORT-base has 67.6\% fewer parameters (3$\times$), ACORT-small has 90.9\% fewer parameters (11$\times$), while ACORT-xsmall has 94.5\% fewer parameters (18$\times$). Even with such large reductions in model sizes, ACORT is still able to provide good performance. Specifically, ACORT-base achieves $+$3.0\% on BLEU-4 and $+$3.7\% on CIDEr, ACORT-small achieves $+$3.8\% on BLEU-4 and $+$3.0\% on CIDEr, whereas ACORT-xsmall achieves $+$1.6\% on BLEU-4 and $+$2.3\% on CIDEr.

ACORT models are also competitive compared to standard SOTA models. The smallest ACORT model mostly outperforms the UD (Up-Down) model, with better scores on BLEU-4, METEOR, ROUGE and CIDEr. Even compared with the best models -- AoANet, M2, and X-Transformer -- all three ACORT models are able to produce competitive metric scores considering their small parameter counts.

In short, ACORT models provide good performance-to-size ratios, demonstrating the effectiveness of the proposed modifications outlined in Sec.~\ref{sec: ACORT: Proposed}.

\subsection{Qualitative Results}
\label{subsec: ACORT: Experiments: Qualitative}

In this section, we present some examples of the generated captions from ACORT along with the baseline ORT-base model. The captions are generated on the validation set using a beam size of 5.

\begin{figure}[t]
    \centering
    \tabulinesep=2pt
    \begin{tabu}{ X X X X }
        \toprule
        \gph{1}{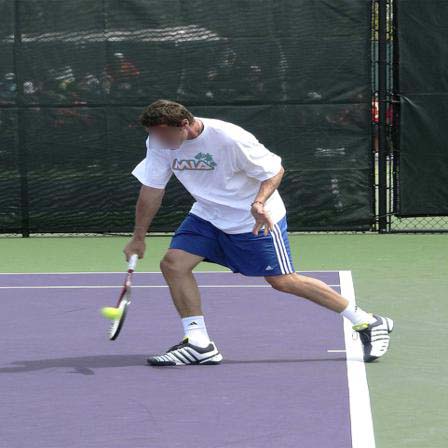} &  
        \gph{1}{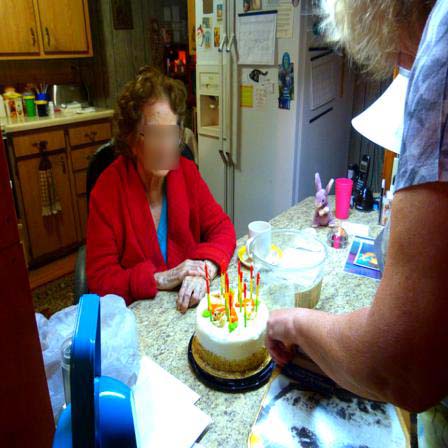} &  
        \gph{1}{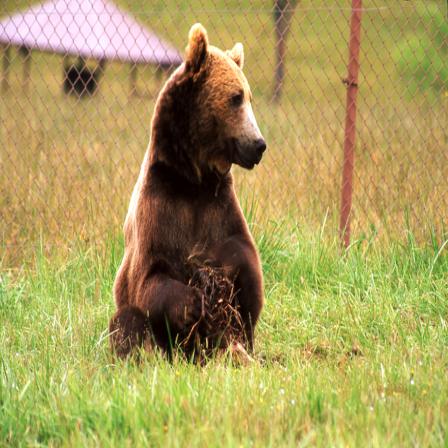} &  
        \gph{1}{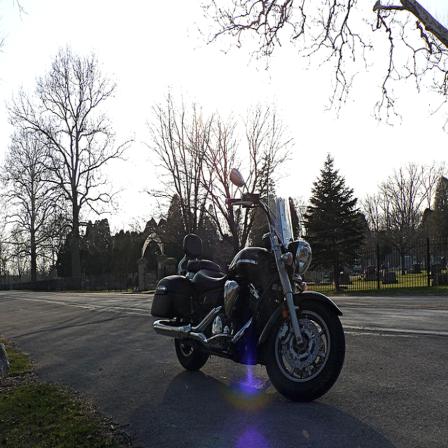} \\
        \tabucline[0.4pt off 2pt]{-}
        \multicolumn{4}{c}{\textit{\textbf{ORT-base}}} \\
        a man standing on a tennis court holding a ball & a woman sitting at a table with a birthday cake with candles on a & a brown bear sitting in the grass                 & two motorcycles parked on the side of a road \\
        \tabucline[0.4pt off 2pt]{-}
        \multicolumn{4}{c}{\textit{\textbf{ACORT-base}}} \\
        a man hitting a tennis ball on a tennis court   & a woman sitting at a table with a birthday cake with candles      & a brown bear sitting in the grass next to a fence & a motorcycle parked on the side of a road    \\
        \tabucline[0.4pt off 2pt]{-}
        \multicolumn{4}{c}{\textit{\textbf{ACORT-small}}} \\
        a man hitting a tennis ball on a tennis court   & a woman sitting at a table with a cake with candles               & a brown bear sitting in the grass next to a fence & a motorcycle parked on the side of a road    \\
        \tabucline[0.4pt off 2pt]{-}
        \multicolumn{4}{c}{\textit{\textbf{ACORT-xsmall}}} \\
        a man hitting a tennis ball on a tennis court   & a woman sitting in front of a birthday cake with candles          & a brown bear sitting in the grass in a field      & a motorcycle parked on the side of a road    \\
        \bottomrule
    \end{tabu}
    \caption{Examples of images where the captions generated by the ACORT models are accurate and descriptive relative to the baseline.}
    \label{fig: ACORT: Qualitative: Model correct}
\end{figure}

\begin{figure}[t]
    \centering
    \tabulinesep=2pt
    \begin{tabu}{ X X X X }
        \toprule
        \gph{1}{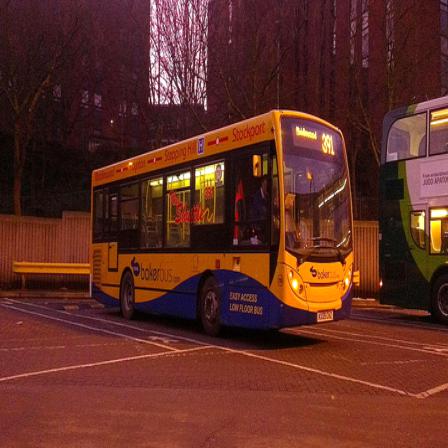} &
        \gph{1}{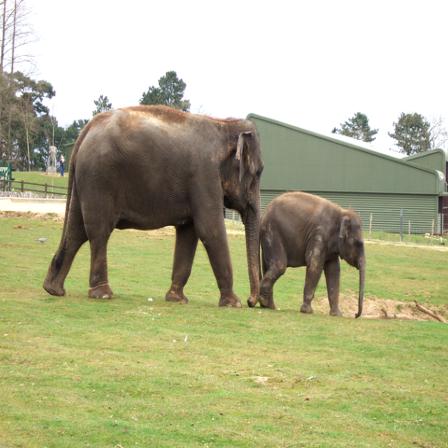} & 
        \gph{1}{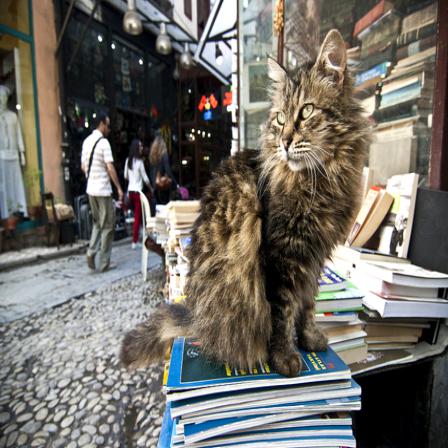} & 
        \gph{1}{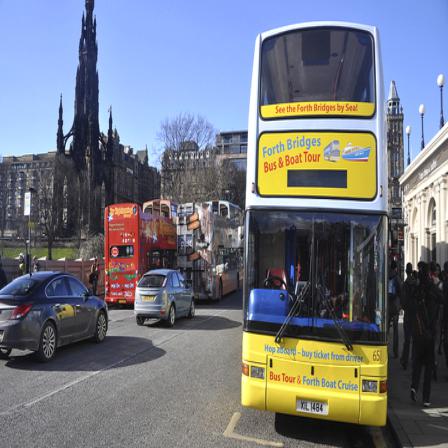} \\
        \tabucline[0.4pt off 2pt]{-}
        \multicolumn{4}{c}{\textit{\textbf{ORT-base}}} \\
        a bus parked in a parking lot at night      & an adult elephant and a baby elephant standing in a field & a cat sitting on top of a pile of books & a yellow double decker bus driving down a street \\
        \tabucline[0.4pt off 2pt]{-}
        \multicolumn{4}{c}{\textit{\textbf{ACORT-base}}} \\
        a blue and yellow bus driving down a street & two elephants standing next to a baby elephant in a field & a cat sitting on top of a book          & a double decker bus driving down a street        \\
        \tabucline[0.4pt off 2pt]{-}
        \multicolumn{4}{c}{\textit{\textbf{ACORT-small}}} \\
        a bus driving down a street at night        & two elephants standing in a field in a                    & a cat sitting on top of a book on a     & a double decker bus driving down a street        \\
        \tabucline[0.4pt off 2pt]{-}
        \multicolumn{4}{c}{\textit{\textbf{ACORT-xsmall}}} \\
        a bus parked in a parking lot at night      & two elephants and a baby elephant standing in a field     & a cat sitting on top of a pile of books & a double decker bus driving down a street with   \\
        \bottomrule
    \end{tabu}
    \caption{Examples of images where the ACORT models made mistakes.}
    \label{fig: ACORT: Qualitative: Baseline correct}
\end{figure}

Fig.~\ref{fig: ACORT: Qualitative: Model correct} showcases some images where the captions generated by the ACORT models are accurate and descriptive relative to the ORT-base model. Despite their small sizes, both ACORT models still managed to convey specific and relevant details regarding the image content. For image (1), ACORT correctly described the action of the player as ``hitting the ball'' rather than holding it. For image (2), all the captions are accurate but ORT-base suffered from ``bad endings'' due to SCST optimisation \cite{guo2018improving}. For image (3), the fence was mentioned by ACORT but ignored by ORT-base. For image (4), the number of motorcycles is correctly specified as one by ACORT instead of two.

Fig.~\ref{fig: ACORT: Qualitative: Baseline correct} showcases images where the captions generated by the ACORT models contain mistakes relative to the ORT-base model. This is to be expected, as the ACORT and baseline models have similar metric scores. From the captions, the presence of bad endings can again be seen. For image (1), even though the bus is correctly described as being ``blue and yellow'' by ACORT, it is incorrectly described as ``driving down a street''; whereas the baseline caption is more accurate. For image (2), ACORT is unable to provide detailed descriptions of the elephants, compared to ORT-base which mentioned both the adult and the baby. Similarly for image (3) and (4), the baseline captions are more detailed, with mentions of ``a pile of books'' and ``a yellow bus''.

Overall, the captions generated by ACORT are still on par with those generated by the baseline ORT model. This shows that caption quality is maintained despite the reductions in parameter count.

\subsection{Ablation Studies}
\label{subsec: ACORT: Experiments: Ablations}

In this section, we provide an extensive analysis of the design choices and architectural modifications made to the baseline ORT model \cite{herdade2019image}. By isolating each of the modifications and comparing them to the baseline, their contribution towards the performance and compactness of the final model ACORT can be better quantified. Unless stated otherwise, any parameter sharing configuration or modification is applied to both the encoder and decoder. All the metric scores presented in this section are obtained using greedy decoding on the MS-COCO validation set.

\subsubsection{Radix encoding}
\label{subsubsec: ACORT: Ablation: Radix encoding}

\begin{table}[t]
    \caption{The effect of different Radix Encoding bases. Overall, Radix Encoding with $v = 768$ provides good performance while maintaining embedding compactness.}
    \label{table: ACORT: Ablation: Radix}
    \centering
    \begin{adjustbox}{max width=\linewidth}
    \begin{tabular}{ `c ~c ~c ~c ~c ~c ~c ~c ~c }
        \toprule
        \fmtr{2.35}{Radix base $v$} & \multicolumn{2}{c}{Params (M)} & \multicolumn{6}{c}{MS-COCO validation scores} \\
                                      \cmidrule(lr){2-3}               \cmidrule(lr){4-9}
                                    & Embeddings       & Total       & B-1   & B-4   & M     & R     & C      & S    \\
        \midrule
        Word (baseline)             & 10.3             & 55.4        & 75.5  & 33.9  & 27.6  & 56.2  & 111.0  & 20.6 \\
        \midrule
        1,024                       & 1.1              & 46.2        & 75.1  & 33.8  & 27.8  & 56.3  & 111.9  & 20.8 \\
        768                         & 0.8              & 46.0        & 75.2  & 33.3  & 27.7  & 55.9  & 111.7  & 20.8 \\
        512                         & 0.5              & 45.7        & 74.9  & 33.6  & 27.9  & 56.2  & 111.9  & 20.7 \\
        256                         & 0.3              & 45.5        & 75.2  & 33.3  & 27.6  & 56.1  & 110.6  & 20.6 \\
        \bottomrule
    \end{tabular}
    \end{adjustbox}
\end{table}

Table~\ref{table: ACORT: Ablation: Radix} demonstrates the validation performance and parameter reductions that can be achieved using Radix Encoding. Across the board from base of $v = 256$ to $v = 1024$, the performance of the models is comparable to that of the baseline. This result shows that the Transformer model can effectively learn the strict dependencies between tokens imposed by Radix Encoding, despite its lack of recurrent connection. 

Furthermore, the sizes of the embedding matrices are reduced significantly up to 97.4\% (10.3M $\rightarrow$ 0.3M). At the same time, the performance degradation at the extreme end is merely 0.36\% for CIDEr (111.0 $\rightarrow$ 110.6). All in all, this result suggests that there exists substantial redundancy in the embeddings matrices, which can be removed without affecting overall performance \cite{shi2018structured,tan2019comic}. The work of \cite{baevski2019adaptive} also found that performing embedding factorisation along with weight sharing can lead to performance improvements in language modelling.

\subsubsection{Cross-layer sharing}
\label{subsubsec: ACORT: Ablation: Layer share}

\begin{table}[t]
    \caption{The effects of varying numbers of independently parameterised layers applied to both the encoder and the decoder. Configuration \texttt{(0x3,1x3)} achieves the closest metric scores when compared to the baseline model while remaining reasonably compact.}
    \label{table: ACORT: Ablation: Layer share}
    \centering
    \begin{adjustbox}{max width=\linewidth}
    \begin{tabular}{ `l ~c ~c ~c ~c ~c ~c ~c ~c }
        \toprule
        \fmtr{2.35}{\makecell{\# independent\\layers}}  & 
        \fmtr{2.35}{\makecell{Layer share\\config}} & 
        \fmtr{2.35}{\makecell{Params\\(M)}} & 
        \multicolumn{6}{c}{MS-COCO validation scores} \\
        \cmidrule(lr){4-9}
        \null               & \null                 & \null     & B-1   & B-4   & M     & R     & C      & S    \\
        \midrule
        6 (baseline)        & (0, 1, 2, 3, 4, 5)    & 55.4      & 75.5  & 33.9  & 27.6  & 56.2  & 111.0  & 20.6 \\
        \midrule
        4                   & (0, 0, 0, 1, 2, 3)    & 40.7      & 75.9  & 33.5  & 27.4  & 55.8  & 110.6  & 20.6 \\
        3 (successive)      & (0, 0, 1, 1, 2, 2)    & 33.4      & 75.9  & 33.9  & 27.5  & 56.2  & 111.0  & 20.5 \\
        3 (symmetric)       & (0, 1, 2, 2, 1, 0)    & 33.4      & 75.8  & 33.3  & 27.0  & 55.6  & 110.5  & 20.3 \\
        2                   & (0x3, 1x3)            & 26.0      & 76.1  & 33.7  & 27.1  & 55.8  & 110.9  & 20.4 \\
        1                   & (0x6)                 & 18.7      & 76.2  & 33.4  & 26.9  & 55.6  & 109.4  & 20.2 \\
        \bottomrule
    \end{tabular}
    \end{adjustbox}
\end{table}

\textbf{Fix the number of layers, vary the number of independent layers:} Table~\ref{table: ACORT: Ablation: Layer share} summarises the performance of baseline networks with varying numbers of uniquely or independently parameterised layers. Cross-layer sharing is able to drastically reduce the number of parameters from 55.4M to just 18.7M, \ie{} a 66.2\% reduction. Furthermore, the performance of the \texttt{(0x6)} model with just one independent layer shared across 6 layers is still comparable to that of baseline, with CIDEr score of 109.4 versus 111.0 and BLEU-1 score of 76.2 versus 75.5. In general, while the broad trend remains that models with more parameters perform slightly better, the parameter-performance trade-off is still favourable to layer-shared models.

\begin{figure}[t]
    \centering
    \begin{subfigure}{.35\linewidth}
        \centering
        \gph{1}{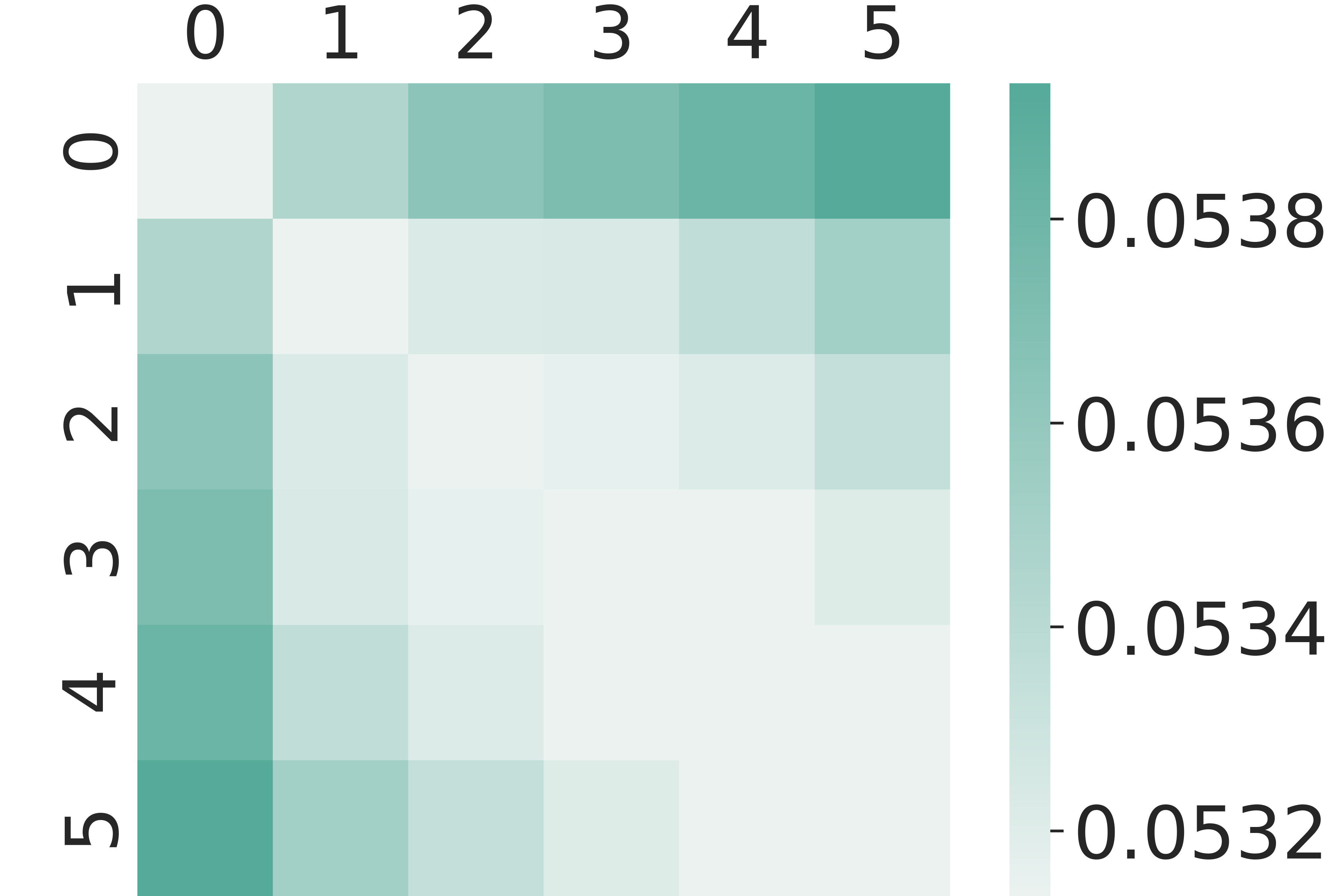}
        \caption{Encoder.}
        \label{fig: ACORT: Layer-sim encoder}
    \end{subfigure}\hspace{0.1\textwidth}%
    \begin{subfigure}{.35\linewidth}
        \centering
        \gph{1}{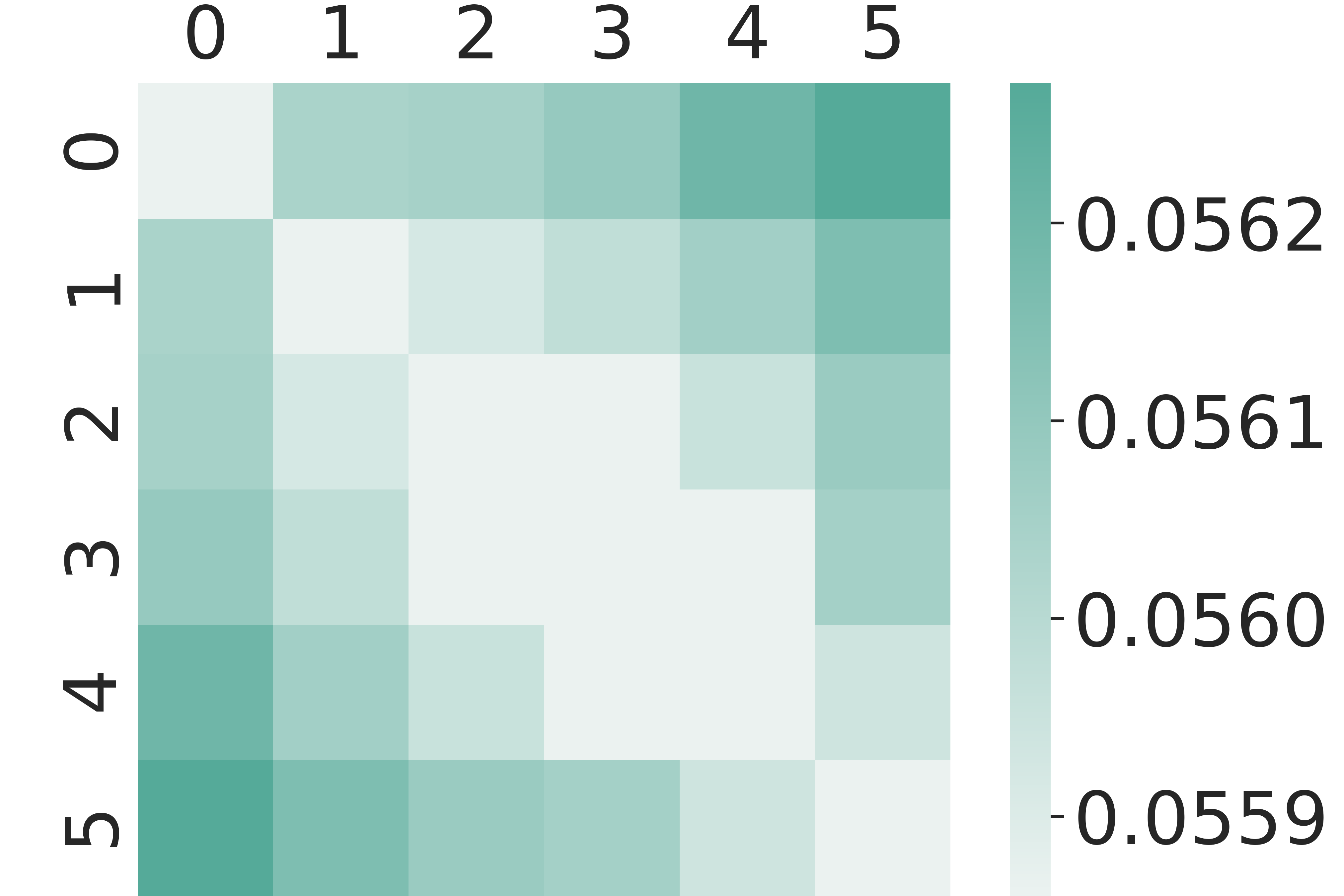}
        \caption{Decoder.}
        \label{fig: ACORT: Layer-sim decoder}
    \end{subfigure}
    \caption{The root mean squared distance for all layer pairs in a standard 6-layer ORT model. Minimum distance in each plot is clipped at its second lowest value. In general, the first layer is more distinct compared to the rest of the layers; layers 2 to 6 are closer to neighbouring layers.}
    \label{fig: ACORT: Layer-sim}
\end{figure}

While Table~\ref{table: ACORT: Ablation: Layer share} provides the performance data of various layer sharing configurations, we seek to explain the reasons behind the performance differences.
To this end, we measured how the parameter distribution of a layer differs from each other in the baseline 6-layer ORT model. The results are given in Fig.~\ref{fig: ACORT: Layer-sim}, obtained by performing $L_2$ normalisation on the last axis\footnote{Normalisation across first or all axes produce similar results.} of the layer parameters, flattening and concatenating them, and computing the pair-wise mean squared distance.

The figure shows that the first layer -- which processes the input embeddings -- is more distinct compared to the rest of the layers. This largely supports the observation that sharing configurations where the first layer is shared more \texttt{(0,0,0,1,2,3)} or shared with the last layer \texttt{(0,1,2,2,1,0)} suffered from worse performance. Configuration \texttt{(0x6)} also suffered from being the worst performer, although that can be attributed in part to its lower parameter count. In addition, the similarity data also show that layers 2 to 6 are closer to neighbouring layers. This suggests that configurations where successive layers are shared will be more performant. Indeed, configurations \texttt{(0,0,1,1,2,2)} and \texttt{(0x3,1x3)} achieve the closest metric scores when compared to the baseline model.

\begin{table}[t]
    \caption{The performance of networks with only 2 independent layers but with varying degree of layer reuse. The 6-layer model with \texttt{(0x3,1x3)} configuration provides the best performance relative to the baseline.}
    \label{table: ACORT: Ablation: Layer reuse}
    \centering
    \begin{adjustbox}{max width=\linewidth}
    \begin{tabular}{ `l ~c ~c ~c ~c ~c ~c ~c ~c }
        \toprule
        \fmtr{2.35}{\# layers}  & 
        \fmtr{2.35}{\makecell{Layer share\\config}} & 
        \fmtr{2.35}{\makecell{Params\\(M)}} & 
        \multicolumn{6}{c}{MS-COCO validation scores} \\
        \cmidrule(lr){4-9}
        \null           & \null                 & \null     & B-1   & B-4   & M     & R     & C      & S    \\
        \midrule
        6 (baseline)    & (0, 1, 2, 3, 4, 5)    & 55.4      & 75.5  & 33.9  & 27.6  & 56.2  & 111.0  & 20.6 \\
        \midrule
        2 (2 ind.)      & (0, 1)                & 26.0      & 75.6  & 33.7  & 27.3  & 55.9  & 110.5  & 20.5 \\
        6 (2 ind.)      & (0x3, 1x3)            & 26.0      & 76.1  & 33.7  & 27.1  & 55.8  & 110.9  & 20.4 \\
        12 (2 ind.)     & (0x6, 1x6)            & 26.0      & 75.9  & 33.4  & 26.9  & 55.7  & 108.9  & 20.2 \\
        \bottomrule
    \end{tabular}
    \end{adjustbox}
\end{table}

\textbf{Fix the number of independent layers, vary the number of layers:} Table~\ref{table: ACORT: Ablation: Layer reuse} demonstrates the performance where the number of independent layers is fixed but each layer is reused across varying numbers of layers in a recurrent fashion similar to \cite{dehghani2019universal}. Here, all the layer-shared models have two independent layers. From the results, even though all three models have very different computation costs -- from 2 layers \texttt{(0,1)} to 12 layers \texttt{(0x6,1x6)} -- they all end up with similar performance. This indicates that repeated computation using the same layer is not an efficient use of compute, with diminishing returns past 3 layer repetitions.
Finally, it can be noted that the 6-layer model with \texttt{(0x3,1x3)} configuration provides the best performance relative to the baseline.

\begin{table}[t]
    \caption{The effects of applying cross-layer sharing to encoder only, decoder only, or both. Sharing both the encoder and the decoder provides the most parameter reduction while maintaining performance.}
    \label{table: ACORT: Ablation: Layer share encoder decoder}
    \centering
    \begin{adjustbox}{max width=\linewidth}
    \begin{tabular}{ `l ~c ~c ~c ~c ~c ~c ~c ~c }
        \toprule
        \fmtr{2.35}{\makecell{\# independent\\layers}}  & 
        \fmtr{2.35}{\makecell{Layer share\\config}} & 
        \fmtr{2.35}{\makecell{Params\\(M)}} & 
        \multicolumn{6}{c}{MS-COCO validation scores} \\
        \cmidrule(lr){4-9}
        \null              & \null              & \null     & B-1  & B-4  & M    & R    & C     & S    \\
        \midrule
        Baseline           & (0, 1, 2, 3, 4, 5) & 55.4      & 75.5 & 33.9 & 27.6 & 56.2 & 111.0 & 20.6 \\
        \midrule
        Share encoder only & (0x6)              & 39.7      & 75.4 & 33.6 & 27.4 & 56.1 & 110.0 & 20.5 \\
        Share decoder only & (0x6)              & 34.4      & 75.8 & 33.5 & 26.9 & 55.6 & 109.0 & 20.4 \\
        Share both         & (0x6)              & 18.7      & 76.2 & 33.4 & 26.9 & 55.6 & 109.4 & 20.2 \\
        \bottomrule
    \end{tabular}
    \end{adjustbox}
\end{table}

\textbf{Apply layer sharing to encoder only, decoder only, or both:} Table~\ref{table: ACORT: Ablation: Layer reuse} shows that applying layer sharing to the encoder only while leaving the decoder intact produced less performance drop compared to sharing the decoder across all metrics except BLEU-1. Interestingly, applying layer sharing to both the encoder and decoder did not seem to worsen overall performance when compared to the decoder-only sharing scheme, with almost no discernible difference in terms of metric scores. However, the number of parameters vastly favoured sharing both the encoder and decoder, with the resulting network having 66\% fewer parameters than the baseline. Overall, sharing both the encoder and the decoder provides the best performance to parameter count trade-off.

\subsubsection{Attention parameter sharing}
\label{subsubsec: ACORT: Ablation: Attention share}

\begin{table}[t]
    \caption{The effect of different attention parameter sharing configurations. Since there seems to be no significant performance difference among the various configurations, \textit{Share-KV} is chosen to allow for key-value reuse.}
    \label{table: ACORT: Ablation: Attention}
    \centering
    \begin{adjustbox}{max width=\linewidth}
    \begin{tabular}{ `l ~c ~c ~c ~c ~c ~c ~c ~c }
        \toprule
        \fmtr{2.35}{\makecell{Attention\\params sharing}} 
        & \multicolumn{2}{c}{Params (M)} & \multicolumn{6}{c}{MS-COCO validation scores} \\
          \cmidrule(lr){2-3}                \cmidrule(lr){4-9}
                                        & Attention        & Total       & B-1   & B-4   & M     & R     & C      & S    \\
        \midrule
        No-Share (baseline)             & 18.9             & 55.4        & 75.5  & 33.9  & 27.6  & 56.2  & 111.0  & 20.6 \\
        \midrule
        Share-KV (encoder)              & 17.3             & 53.9        & 75.6  & 33.9  & 27.5  & 56.1  & 111.0  & 20.4 \\
        Share-KV (decoder)              & 15.8             & 52.3        & 75.6  & 33.9  & 27.4  & 56.1  & 111.1  & 20.7 \\
        Share-KV                        & 14.2             & 50.7        & 75.6  & 33.8  & 27.6  & 56.2  & 111.2  & 20.7 \\
        Share-QK                        & 14.2             & 50.7        & 75.7  & 34.0  & 27.7  & 56.3  & 111.9  & 20.7 \\
        \bottomrule
    \end{tabular}
    \end{adjustbox}
\end{table}

Table~\ref{table: ACORT: Ablation: Attention} shows the impact of various parameter sharing configurations. All the attention-shared models are performing surprisingly well despite the reduction in attention parameters. This is especially true for the \textit{Share-QK} model, as it is able to outperform the baseline model across all metrics. We hypothesise that this is due to the increased regularisation from weight sharing. While the \textit{Share-KV} model performs slightly inferior to \textit{Share-QK}, it has a minor advantage in terms of compute cost (as mentioned in Sec.~\ref{subsec: ACORT: Proposed: Attention}). Finally, there seems to be no significant performance difference whether the sharing is applied to encoder only, decoder only, or to both. It can be noted that sharing the decoder's attention parameters provides larger parameter reduction than the encoder, as the decoder contains cross-attention layers absent from the encoder.


\section{Limitations and Future Work}
\label{sec: ACORT: Limitations}

One of the limitations of this work is the number of hyperparameters introduced. In particular, cross-layer parameter sharing introduced a wide range of possible layer-sharing configurations. While traversing all the possible hyperparameter combinations can be expensive, we believe the comprehensive baseline comparison and ablation study provided in Sec.~\ref{subsec: ACORT: Experiments: Baseline} and \ref{subsec: ACORT: Experiments: Ablations} can mitigate this shortcoming partially. Alternatively, efficient hyperparameter optimisation (HPO) methods \cite{lorraine2020optimizing,dong2021autohas} can be used to further search the hyperparameter space. However, since this work focuses on demonstrating the effectiveness of the proposed parameter reduction techniques rather than pursuing raw performance, we leave the use of HPO techniques to future work.

Besides that, the Radix Encoding method utilised performs token factorisation across time steps, leading to longer sequence lengths. In the future, we wish to incorporate additional neural encoding layers to compress and consolidate tokens across time steps. Alternatively, an ALBERT-style token factorisation could be used. At the output layer, strategies such as \cite{chen2016strategies} and CNN-softmax \cite{jozefowicz2016exploring} can be used to achieve model size reduction.

On the other hand, while the various parameter reduction techniques presented in this work are only applied to a Transformer-based model (ORT), all of them are in principle compatible with other types of image captioning models. For example, cross-layer parameter sharing can potentially be applied to a multi-layer Recurrent Neural Network (RNN) decoder \cite{anderson2018bottom} in order to share weights across the RNN layers. Such decoder can also be outfitted with the \textit{Share-KV} cross-attention module as described in Sec.~\ref{subsec: ACORT: Proposed: Attention}. Finally, Radix Encoding can and has been successfully used together with an RNN-based model \cite{tan2019comic}. Similarly, models with convolutional decoder \cite{li2020dual} can employ cross-layer parameter sharing to share weights across convolutional layers, while simultaneously utilising \textit{Share-KV} cross-attention and Radix Encoding. The exploration of these architectures is left as future work.


\section{Conclusion}
\label{sec: ACORT: Conclusion}

We have presented ACORT -- A Compact Object Relation Transformer architecture for parameter efficient image captioning. ACORT models can achieve performances that are on par with the standard ORT model (CIDEr score $\geq$126) but with 3.7$\times$ to 21.6$\times$ fewer parameters. This is achieved via the incorporation of three parameter reduction methods: Radix Encoding, cross-layer parameter sharing, and attention parameter sharing. Our results on the MS-COCO dataset demonstrates that the proposed ACORT-base and ACORT-small models are capable of achieving metric scores that are competitive against baselines and SOTA approaches. Finally, qualitative results and ablation studies are also presented to further demonstrate the effectiveness of the proposed modifications. Together with the public release of model code and pre-trained checkpoints, we hope that these findings can spur research interest in the image captioning community.


\section*{Acknowledgement}
This research is supported by the Fundamental Research Grant Scheme (FRGS) MoHE Grant FP021-2018A, from the Ministry of Education Malaysia.


\bibliographystyle{elsarticle-num}
\bibliography{ref}

\end{document}